\documentclass[conference]{IEEEtran}
\IEEEoverridecommandlockouts

\usepackage{cite}
\usepackage{amsmath,amssymb,amsfonts}
\usepackage{algorithmic}
\usepackage{graphicx}
\usepackage{textcomp}   
\usepackage{comment}    
\usepackage{xcolor}     
\usepackage{tabularx}   
\usepackage{fancybox,framed}    
\usepackage{listings}   
\usepackage{url}        
\usepackage{booktabs}   
\usepackage{microtype}  
\usepackage{hyperref}
\usepackage{cleveref}   

\usepackage{sourcecodepro}
\usepackage[T1]{fontenc}

\def\BibTeX{{\rm B\kern-.05em{\sc i\kern-.025em b}\kern-.08em
    T\kern-.1667em\lower.7ex\hbox{E}\kern-.125emX}}
\begin{document}

\lstset{
  basicstyle=\footnotesize\ttfamily,
  breaklines=true,
  breakatwhitespace=true,
}

\def\citepunct{, }
\def\citedash{--}

\newcommand{\todo}[1]{\textbf{\color{red}[TODO: #1]}}
\AtBeginDocument{%
  \providecommand\BibTeX{{%
    Bib\TeX}}}

\title{Toxicity in Twitch Chats: An LLM-Based Analysis Across Gaming Communities

}

\makeatletter
\newcommand{\linebreakand}{%
  \end{@IEEEauthorhalign}
  \hfill\mbox{}\par
  \mbox{}\hfill\begin{@IEEEauthorhalign}
}
\makeatother

\author{
  \IEEEauthorblockN{Ronja Fuchs}
  \IEEEauthorblockA{\textit{Institute for Information Processing}\\
  \textit{Leibniz University Hannover}\\
  Hannover, Germany \\
  fuchsron@tnt.uni-hannover.de}
  \and
  \IEEEauthorblockN{Florian Rupp}
  \IEEEauthorblockA{\textit{Department of Computer Science}\\
  \textit{Technische Hochschule Mannheim}\\
  Mannheim, Germany \\
  florian.rupp95@gmail.com}
  \and
  \IEEEauthorblockN{Timo Bertram}
  \IEEEauthorblockA{\textit{Institute for Machine Learning}\\
  \textit{Johannes Kepler University}\\
  Linz, Austria\\
  bertram@ml.jku.at}
  \linebreakand
  \IEEEauthorblockN{Kai Eckert}
  \IEEEauthorblockA{\textit{Department of Computer Science}\\
  \textit{Technische Hochschule Mannheim}\\
  Mannheim, Germany \\
  k.eckert@hs-mannheim.de}
  \and
  \IEEEauthorblockN{Alexander Dockhorn}
  \IEEEauthorblockA{\textit{SDU Metaverse Lab}\\
  \textit{University of Southern Denmark}\\
  Odense, Denmark \\
  adoc@sdu.dk}
}

\IEEEoverridecommandlockouts
\IEEEpubid{\makebox[\columnwidth]{ 979-8-3315-9476-3/26/\$31.00 \textcopyright2026 IEEE\hfill} 
\hspace{\columnsep}\makebox[\columnwidth]{ }}

\maketitle

\IEEEpubidadjcol

\begin{abstract}
Toxicity in online gaming communities remains a persistent challenge, manifesting across genres, platforms, and player interactions. While much research is focused on in-game toxicity, less is known about how toxic behavior varies between gaming communities on streaming platforms. 
To address this shortcoming, we analyze approximately 20 million chat messages from 4,452 streams, spanning seven game genres on Twitch. 
We categorize messages according to Twitch's toxicity taxonomy with a pre-trained Large Language Model using zero-shot classification. The taxonomy comprises four categories and eight subclasses, including harassment, discrimination, sexual content, and profanity. 
Our approach achieves an F1 score of 94.5\% on the TextDetox dataset and demonstrates human-model agreement comparable to inter-human agreement. 
Our analysis reveals that 2.4\% of all messages are classified as toxic, with notable differences across genres: streams of MOBA games exhibit the highest relative rate of toxicity (3.2\%), and sports games show the lowest rate (2\%). 
Furthermore, results indicate that individual games differ significantly in their toxicity distributions, even within genres, suggesting the existence of game-specific community norms and mechanics that shape toxic behavior beyond genre-level effects. 
These findings offer empirical insights into genre- and game-specific toxicity patterns on Twitch and can inform more targeted moderation strategies for gaming communities.
\end{abstract}


\section{Introduction}

Online gaming spaces have become important social environments where players and viewers interact, form communities, and establish shared norms. However, these environments are often shaped by harmful behavior, including harassment, abusive language, and discrimination. Such behavior can negatively affect player well-being, discourage participation, and undermine the long-term health of gaming communities~\cite{FROMMEL2023100302, fuchs2025players, fox2017women, ma2024gaming}.
Although widely recognized by developers, researchers, and players, the problem remains persistent and difficult to address~\cite{wijkstra2023help}. One reason is that harmful behavior takes many forms, and perceptions of toxicity are subjective and context-dependent~\cite{frommel2022challenges}.
Most current interventions respond only after harmful interactions have already occurred~\cite{wijkstra2023help}.
Yet such experiences can cause players to withdraw from games or gaming communities entirely~\cite{fuchs2025players, fox2017women}, and may even discourage new players from engaging with games in the first place~\cite{martens2015toxicity}.
Game-external communities play an important role in gaming by providing social spaces that motivate players and can positively influence their mental well-being~\cite{ma2024gaming}.



Twitch\footnote{\url{https://www.twitch.tv/}; Last accessed: \today} is a popular streaming platform where players broadcast gameplay to live audiences, creating shared social environments in which viewers interact through real-time chat. While this fosters community engagement, it can also enable toxic exchanges. 
Prior work has identified Twitch as a platform in which toxic communication is prevalent~\cite{gandolfi2022sharing}. Unlike in-game environments, where harmful interactions are often tied to player performance, toxic behavior on Twitch emerges more strongly from the social dynamics of chat.
Although Twitch offers automatic moderation tools, moderation practices largely remain the responsibility of streamers.
To support moderation and toxicity prevention, it is important to understand how toxic behavior manifests across different genres, individual games, and Twitch communities. Identifying such patterns can provide insights into how toxicity varies across environments and help inform more targeted moderation strategies.

In this work, we analyze toxic behavior in Twitch text chats across different gaming communities, examining how chat interactions vary between game genres and how these differences appear in patterns of toxicity.
We compile a corpus of approximately 20 million chat messages of the two most popular games across seven selected Twitch categories and classify them into Twitch-defined toxicity types using zero-shot classification with a pre-trained Large Language Model (LLM).
This work provides a reproducible pipeline for toxicity detection and classification of text messages, along with an analysis of toxicity in Twitch chats across multiple popular genres and streamers. Our findings offer insights into harmful dynamics within Twitch communities and may support improved toxicity detection and prevention systems.

The remainder of this paper is structured as follows: Section~\ref{sec:background} reviews related work on toxicity detection, including studies using LLMs and research on Twitch. Section~\ref{sec:methodology} describes the methodology, and Section~\ref{sec:results} presents the results. Section~\ref{sec:discussion} discusses the findings and limitations, followed by the conclusion in Section~\ref{sec:conclusion}.

%

\section{Background and Related Work}
\label{sec:background}

\subsection{Defining Toxicity}
Prior work highlights that toxicity, hate speech, and abusive behavior are inherently context-dependent and partly subjective, which leads to varying definitions and annotation schemes across studies~\cite{barbarestani-etal-2024-content, fox2017women, gandolfi2022sharing}.
Twitch itself defines four categories of inappropriate or harmful messages with eight subclasses in their automatic moderation (cf. Table~\ref{tab:twitch-classification}).\footnote{\url{https://help.twitch.tv/s/article/how-to-use-automod}; \\\hphantom{ww}Last accessed: \today} As this work focuses on data from Twitch, we use this categorization and definition for our evaluation.

\begin{table}[t]
\centering
\caption{Categories and Subclasses of Inappropriate or Harmful Messages as given by Twitch}
\label{tab:twitch-classification}
\begin{tabularx}{\columnwidth}{lX}
\toprule
\textbf{Category} & \textbf{Description} \\
\midrule
\multicolumn{2}{l}{\textbf{Harassment}} \\
Aggression & Threatening, inciting, or promoting violence or other harm \\
Bullying & Name-calling, insults, or antagonization \\
\midrule
\multicolumn{2}{l}{\textbf{Discrimination and Slurs}} \\
Disability & Demonstrating hatred or prejudice based on perceived or actual mental or physical abilities \\
Sexuality, sex, or gender & Demonstrating hatred or prejudice based on sexual identity, sexual orientation, gender identity, or gender expression \\
Misogyny & Demonstrating hatred or prejudice against women, including sexual objectification \\
Race, ethnicity, or religion & Demonstrating hatred or prejudice based on race, ethnicity, or religion \\
\midrule
\multicolumn{2}{l}{\textbf{Sexual Content}} \\
Sex-based terms & Sexual acts, anatomy \\
\midrule
\multicolumn{2}{l}{\textbf{Profanity}} \\
Swearing & Swear words, \&\^{}\#\$\%\* \\
\bottomrule
\end{tabularx}
\end{table}

\subsection{Toxicity Detection with Large Language Models}

As widely accessible tools for data processing, LLMs have been applied to a variety of text-based datasets, including those used for toxicity detection~\cite{albladi2025toxicLLM, dementieva2024overview, dementieva2025overview}. 

Some studies aim to improve toxicity detection through approaches such as knowledge graphs~\cite{zhao2025enhancing}, distilled models~\cite{zhang2024efficient}, or models specifically fine-tuned for toxicity-related tasks~\cite{dementieva2025overview}. Other work has evaluated the accuracy of LLMs in detecting toxic content~\cite{kruschwitz2024llm, koh2024can}. In addition, LLMs have been proposed as automated moderation tools in gaming contexts~\cite{yang2025unified}. Overall, prior research suggests that LLMs achieve promising performance in toxicity detection~\cite{gretz_zero-shot_2023,savelka_unreasonable_2023}.

\subsection{Toxicity Detection and Moderation on Twitch}

Twitch tackles the problem of toxicity with two methods, automatic (AI-based) and manual moderation.
Gandolfi and Ferdig~\cite{gandolfi2022sharing} emphasize the role of Twitch as a bearer of toxicity with few opportunities for contradiction. They furthermore analyze data from Dota 2 streams on Twitch.

Kim et al.~\cite{kim2022understanding} show that Twitch users often bypass automated moderation by replacing text with emotes, which humans can interpret but systems struggle to detect. They compiled a dataset of Twitch emotes and developed a visual classifier, finding that approximately 3.82\% of Twitch chat messages are toxic.

Huth et al.~\cite{huth2025exploring} present a real-time toxicity detection plugin for Twitch using the Google Perspective API.\footnote{\url{https://www.perspectiveapi.com/}
; Last accessed: \today} They highlight moderation challenges, particularly for smaller streamers. Although designed for Twitch chat, their pipeline was evaluated using the Jigsaw Toxic Comment Classification dataset.\footnote{\url{https://www.kaggle.com/c/jigsaw-toxic-comment-classification-challenge}; \\\hphantom{ww}Last accessed: \today}

Dreier and Pirker~\cite{dreier2023toxicity} analyze toxicity across 36 Twitch channels and 100,000 messages, considering stream type, streamer gender, community size, and game genre. They find higher toxicity in multiplayer streams than in single-player ones and slightly more in shooter games. Smaller streams often lack moderation resources, while larger audiences are associated with more hate messages toward both viewers and streamers.

\section{Methodology}
\label{sec:methodology}

\subsection{Research Questions}
This paper analyzes toxic behavior in Twitch chat with a focus on community-level differences. We aim to understand how different communities behave while watching popular streams and to identify potential differences between them.
To address our research objective using appropriate statistical measures and ensure transparency, we apply the Goal–Question–Metric (GQM) as proposed by Wohlin et al.~\cite{wohlin2012experimentation} and formulate the following research questions:

\begin{figure}[t]
    \centering
    \includegraphics[width=0.5\textwidth]{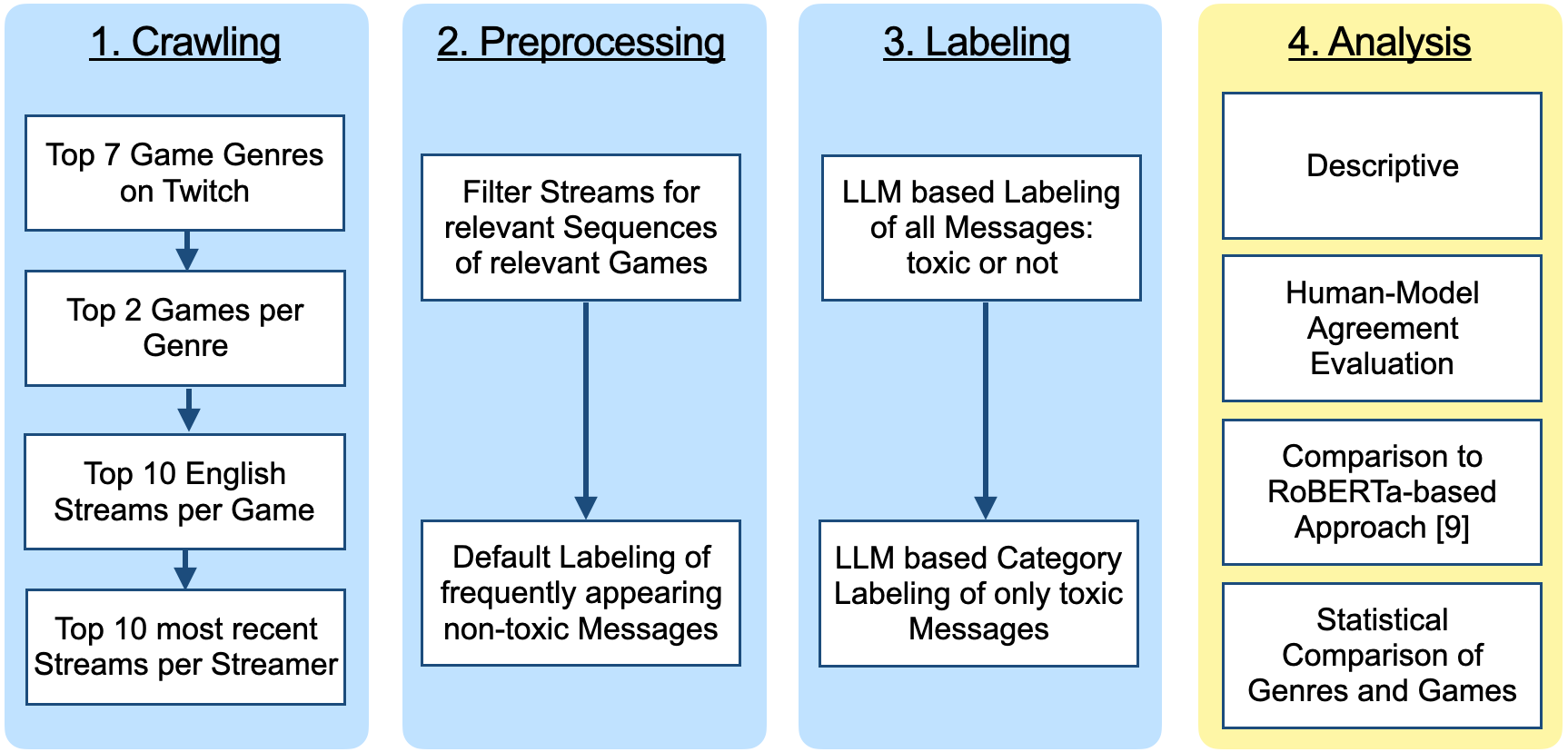}
    \caption{Our pipeline for toxicity detection follows four sequential steps: data crawling, data preprocessing, LLM-based labeling, and data analysis.}
    \label{fig:process}
\end{figure}

\begin{framed}
    \begin{enumerate}
        \item[\textbf{RQ1}] What forms of toxic behavior are most commonly exhibited by viewers on Twitch?
        \item[\textbf{RQ2}] How does viewer toxicity on Twitch vary between games?
        \item[\textbf{RQ3}] How does viewer toxicity on Twitch vary across genres?
    \end{enumerate}
\end{framed}

To address these questions, we use three sets of metrics: 
\begin{enumerate}
    \item[\textbf{M1.1}] Ratio of toxic messages, overall and with regard to toxicity categories and subclasses.
    \item[\textbf{M1.2}] Frequency of co-occurrences of toxicity subclasses.
    \vspace{1em}
    \item[\textbf{M2.1}] Ratio of toxic messages in games, overall and within toxicity categories and subclasses.
    \item[\textbf{M2.2}] Comparison of distributions of toxicity subclasses between games.
    \vspace{1em}
    \item[\textbf{M3.1}] Ratio of toxic messages in genres, overall and within toxicity categories and subclasses.
    \item[\textbf{M3.2}] Comparison of distributions of toxicity subclasses between genres.
\end{enumerate}

\subsection{Toxicity Category Selection}
\label{subsec:toxic_cat_selection}

For the classification of toxic chat messages, we adopt the toxicity taxonomy used by Twitch (cf. Table~\ref{tab:twitch-classification}). We use this taxonomy for two reasons: (1) there is no single universally accepted categorization of toxicity in the literature, and (2) all messages analyzed in this work originate from Twitch chat. We rely on the platform-specific moderation framework that governs this communication space. 
By using Twitch’s moderation taxonomy, we ensure that our classification scheme is closely aligned with the context in which the messages were produced.

\subsection{Twitch Game Categories and Genre Selection}

To create our data set, we selected two games from several Twitch categories with high viewer representation on the platform. Table~\ref{tab:twitch-categories} shows the selected categories and corresponding games based on their popularity on Twitch as of 14 November 2025.

Because genre boundaries in games are often blurred, we use Twitch’s category system as a practical basis for selection. At the same time, Twitch categories do not always group games with comparable gameplay or genre characteristics.
As a result, Twitch's category labels may result in substantial internal variation among games covered by the same label. 
We alter the given categorization such that categories only contain games with comparable dynamics. First, we separate \textit{Strategy} into \textit{Multiplayer (MP) Strategy} and \textit{Single-Player (SP) Strategy}, with the same reasoning as the existing separation of \textit{Multiplayer Shooters} and \textit{Single-Player Shooters}: although both fall under a broader genre umbrella, they differ in gameplay structure, pacing, and player interaction, all of which may also shape chat behavior. 

All selected games contribute to the analyses for \textbf{RQ1} and \textbf{RQ2}, but only a subset of categories is used for the genre-focused analysis in \textbf{RQ3}.
As the strategy-related categories and the MMO category are less internally homogeneous, we exclude them from the genre-focused comparison.
For the remainder of this paper, the term \textit{genre} is only used when referring to these four selected categories, while \textit{game category} denotes Twitch game categories more generally.

\begin{table}[t]
\centering
\caption{Our selection of the two most popular games per category on Twitch.}
\label{tab:twitch-categories}
\resizebox{1.0\columnwidth}{!}{
    \begin{tabular}{@{}llll@{}}
        \toprule\textbf{Twitch Game Category} & \textbf{Genre} & \textbf{Game} 1 & \textbf{Game 2} \\
        \midrule
        MOBA & MOBA & League of Legends & Dota 2 \\
        FPS  & MP Shooter & Counter-Strike~2 & Valorant \\
        Shooter & SP Shooter & Cyberpunk 2077 & Red Dead Redemption 2 \\
        Sports Games & Sports Games & FIFA/FC 26 & Trackmania \\
        \midrule
        MMO & --- & Path of Exile & Minecraft \\
        \lbrack MP\rbrack~ Strategy & --- & Dead by Daylight & Hearthstone \\
        \lbrack SP\rbrack~ Strategy & --- & Dispatch & Plants vs Zombies \\
    \bottomrule
    \end{tabular}
}
\end{table}
\subsection{Data Set Generation and Preprocessing}
\label{sec:data-preprocessing}

Using data from SullyGnome,\footnote{\url{https://sullygnome.com/}; Last accessed: \today} a website that displays Twitch statistics and analysis, we selected the 50 most-viewed streamers per game. 
For each of these streamers, we downloaded the chats of their ten most recent Videos-on-Demand (VODs) using the TwitchDownloaderCLI.\footnote{Provided by: \url{https://github.com/lay295/TwitchDownloader}, \\\hphantom{ww}Last accessed: \today}
This results in a dataset with a total of 4,452 streams with 20,212,682 messages and 29,001 hours of stream time from the time span of 19 June 2024 to 14 November 2025. Labeling all messages required approximately 306 hours of compute time on four A6000 GPUs.

From the approximately 20 million chat messages, 34.7\% consist only of one word.
To reduce computational effort for the classification of each message, we pre-label objectively non-toxic messages such as ''hi'', ''yes'', or ''gg'' from the set of the 50 most occurring messages. 
A full list of pre-labeled messages is available in our supplementary material~\cite{supplementary}.
Twitch allows streamers to manually set up bots that automatically give predefined information to viewers.
We pre-label messages from known bots, such as ''Nightbot'' or ''StreamElements'', accordingly.

\begin{table*}[t]
\centering
\caption{Exemplary results for each toxicity category (see Table~\ref{tab:twitch-classification}). Two messages are shown per category for clarity.}
\label{tab:examples}
\begin{tabular}{@{}lll@{}}
\toprule\textbf{Toxicity Class} & \textbf{Example 1} & \textbf{Example 2} \\
\midrule
Harassment Aggression & <user> <- he deserves worse than just a shock 
 &  MODS chop off his cock \\
Harassment Bullying & STOP CHEATING IDIOTS & bro you are the most useless supp ever \\
Discrimination Disability & he has autism for sure & is she retarded \\
Discrimination Sexuality/Gender & Gayge & they all egirls \\
Discrimination Misogyny & women = annyoing voice = skip [...] & its a shame she aint hot \\
Discrimination Race/Religion & nigga flooding the chat with bullshit lmaooo
 & you are redneck white \\
Sexual Content & deal I would suck you off then & unbutton your pants just a lil bit \\
Profanity Targeted & ASS & fu ck  \\
\bottomrule
\end{tabular}
\end{table*}

\subsection{Zero Shot Classification: Toxic vs. Non-toxic}

To automatically label whether a chat message is toxic, we use a zero-shot classification approach with an LLM and an instruction prompt. 
Because the interpretation of toxicity often depends on conversational context, we provide the model with a temporal context window of the preceding ten seconds of chat messages to support more informed judgments. 

After preliminary experiments, we selected Phi4~\cite{abdin_phi-4_2024}, a 14B-parameter model, for this task. Phi4 is suitable for toxicity detection because, unlike models such as Llama3~\cite{grattafiori_llama_2024}, it is less optimized to refuse or filter abusive content.
Its 16k token context window enables nuanced language understanding while maintaining relatively fast inference times. Larger models may offer slight accuracy improvements but require substantially greater computational resources.

In the first step, we prompt the LLM as a binary toxicity classifier using the label definitions from Table~\ref{tab:twitch-classification}. Next, messages classified as toxic are labeled using a modified prompt based on the same toxicity definitions. Here, the model assigns each message to a specific toxicity class. We provide both prompts in the supplementary material~\cite{supplementary}.

\section{Results}
\label{sec:results}

Of our dataset of 20,212,682 messages, 14.4\% were pre-labeled as not toxic by default (cf. Section~\ref{sec:data-preprocessing}) and 2.4\% of all messages were labeled as toxic by our system, approximately matching results from previous studies~\cite{kim2022understanding}.
Table~\ref{tab:examples} presents exemplary messages classified as toxic, with two examples shown per toxicity category and subclass.
During the binary labeling process, a very small proportion (0.06\%) was incorrectly labeled by the LLM, producing outputs that did not conform to the binary yes/no label definition. Due to their small proportion, these cases are excluded from subsequent analyses.


Before analyzing and discussing the results in more detail, we evaluate the labeling quality through a small-scale agreement study and a comparison with an existing toxicity dataset containing known labels.

\subsection{Human-Model Agreement Evaluation}
\label{sec:model_agreement}
Human-model agreement was tested on 100 messages (50 toxic, 50 non-toxic) with ten seconds context; three game researchers (2M, 1F) independently labeled for toxicity, blinded.

Agreement was evaluated using Cohen’s Kappa~\cite{cohen1960coefficient} (\mbox{-1 = disagreement}, 1 = perfect agreement), achieving an average Kappa of 0.53 (0.44, 0.46, 0.70), which indicates moderate agreement~\cite{landis1977measurement}. Human raters reached a similar score of 0.55 (0.48, 0.51, 0.66), suggesting comparable human–model and inter-human agreement. This indicates the labeling is sufficiently reliable for further analysis.

Human–model agreement for the eight toxicity subclasses was evaluated using the same sample of toxic messages. Human raters assigned subclass labels, resulting in an average Cohen’s Kappa of 0.42 (0.24, 0.33, 0.70) and inter-human agreement of 0.32 (0.18, 0.26, 0.52). Although lower, these values still indicate moderate agreement.
A large proportion of labeling disagreements (30.6\% of all disagreements) occurred when the LLM assigned \textit{bullying} while human raters assigned \textit{profanity}. Another frequent mismatch was between \textit{aggression} assigned by the LLM and \textit{bullying} assigned by human raters (16.1\%).

\subsection{Comparison to a fine-tuned BERT-based approach}

We further compare our zero-shot toxicity detection approach with the fine-tuned RoBERTa model proposed by Dementieva et al.~\cite{dementieva2025overview}.
Applying our method to the English subset of the TextDetox dataset (5,000 labeled messages)\footnote{\url{https://huggingface.co/textdetox/xlmr-large-toxicity-classifier-v2};\\\hphantom{ww}Last accessed: \today} yields an F1 score of 94.5\%, slightly exceeding the 92.3\% reported by Dementieva et al.~\cite{dementieva2025overview}.
Although the dataset contains general language without conversational context, this result indicates that the approach generalizes well beyond the Twitch domain.

\subsection{RQ1 Results: Commonly exhibited toxic behavior on Twitch}

Overall, 2.39\% (472{,}891) of messages were labeled as toxic, while of those, 95.8\% have both a primary and a secondary label assigned.
Results show that the most toxic messages were labeled as \textit{harassment} (primary label/secondary label: 75.6\% / 12.7\%), followed by \textit{discrimination} (12.0\% / 9.5\%), \textit{profanity} (10.0\% / 53.0\%), and \textit{sexual content} (2.4\% / 4.2\%). 
The most common subclasses within toxic messages were \textit{bullying} (primary label/secondary label: 61.0\% / 11.0\%) and \textit{aggression} (14.6\% / 1.8\%), followed by \textit{profanity} (10.0\% / 53.0\%). These are followed by discrimination based on \textit{race, ethnicity, or religion} (5.3\% / 1.9\%), \textit{sexuality or gender} (3.8\% / 2.5\%), \textit{sexual content} (2.4\% / 4.2\%), \textit{misogyny} (2.0\% / 3.3\%), and discrimination based on \textit{disability} (0.9\% / 1.9\%).

The most frequent co-occurrences between primary and secondary labels are \textit{harassment} with \textit{profanity} (54.0\% of toxic messages), followed by \textit{discrimination} with \textit{harassment} (10.8\%) and \textit{harassment} with \textit{sexual content} (3.0\%).
The most common co-occurrences for subclasses were \textit{bullying} with \textit{profanity} (43.0\%), \textit{aggression} with \textit{profanity} (10.9\%), and discrimination based on \textit{race, ethnicity, or religion} with \textit{bullying} (3.2\%). Aggregating subclasses into the broader categories shown in Table~\ref{tab:twitch-classification} reveals that 85.6\% of toxic messages contain \textit{harassment}, 63.0\% \textit{profanity}, 18.0\% \textit{discrimination}, and 6.6\% \textit{sexual content}. 

Streams were further grouped into high- and low-toxicity based on their mean toxicity rate. After testing for variance differences with Levene's test and applying ANOVA or Welch's t-test where appropriate, significant differences emerged for three out of seven subclasses: \textit{sexual content} (\mbox{$F=44.12$}, \mbox{$p<0.001$}), discrimination based on \textit{sexuality or gender} (\mbox{$t=14.56$}, \mbox{$p<0.001$}), and \textit{aggression} (\mbox{$t=18.66$}, \mbox{$p<0.001$}), with only the latter occurring more frequently in low-toxicity streams.




\subsection{RQ2 Results: Variations of viewer toxicity between games}
The most prevalent game was Counter-Strike~2, making up 28\% of all messages in the dataset, followed by League of Legends (15.5\%), Valorant (12.9\%), and Path of Exile (11.5\%). Minecraft held the fewest messages (0.07\%). 

The relative amount of toxic messages per game is displayed in Table~\ref{tab:relative_toxicity_per_game}, with Red Dead Redemption II being the most toxic (4.0\%) and Minecraft being the least toxic (0.7\%).
The distribution of toxicity categories as primary label per game is displayed in Figure~\ref{fig:toxic_cat_game}.
Since \textit{harassment} and \textit{discrimination} include multiple subclasses, we introduced a binary indicator marking whether any subclass of these categories appeared in a message. 
Across all games, \textit{harassment} was by far the most common primary toxicity category (67.8\%-89\%).
\textit{Discrimination} ranked second (5.0\%-16.4\%), with \textit{sexual content} (2.0\%-4.2\%) ranking last, except in EA Sports FC 26 where \textit{profanity} ranked second (10.0\%) and \textit{discrimination} third (9.0\%).
For secondary labels, the order was consistent across all games: \textit{profanity} (56.6\%-79\%) was by far the most prevalent, \textit{harassment} (10.5\%-19.5\%), \textit{discrimination} (8.1\%-15.2\%), and \textit{sexual content} (2.3\%-9.6\%) followed after.
At the subclass level, \textit{bullying} (9.3\%-16.9\%), \textit{aggression} (1.2\%-2.9\%), and \textit{profanity} (4.0\%-11.6\%) were the most common primary labels. The least frequent subclass was typically discrimination based on \textit{disability} (0.4\%-1.2\%), except in Minecraft (1\%) and League of Legends (1.8\%), where discrimination based on \textit{sexuality or gender} and \textit{misogyny}, respectively, occurred least often. For secondary labels, \textit{profanity} (56.6\%-79.1\%) was again most prevalent, followed by \textit{bullying} (9.3\%-16.9\%).

We compared per-stream distributions of toxicity subclasses in primary labels using PERMANOVA~\cite{anderson2014permutational}. Pairwise comparisons showed significant differences between most games, except for Counter-Strike~2 and Valorant, and Plants vs. Zombies and Trackmania. For the latter pair, PERMDISP~\cite{anderson2006distance} indicated unequal dispersion, suggesting the PERMANOVA result may reflect differences in variability rather than distribution centers.

\begin{figure}[t]
    \centering
    \includegraphics[clip, trim=0cm 16cm 0cm 0cm,width=0.98\columnwidth]{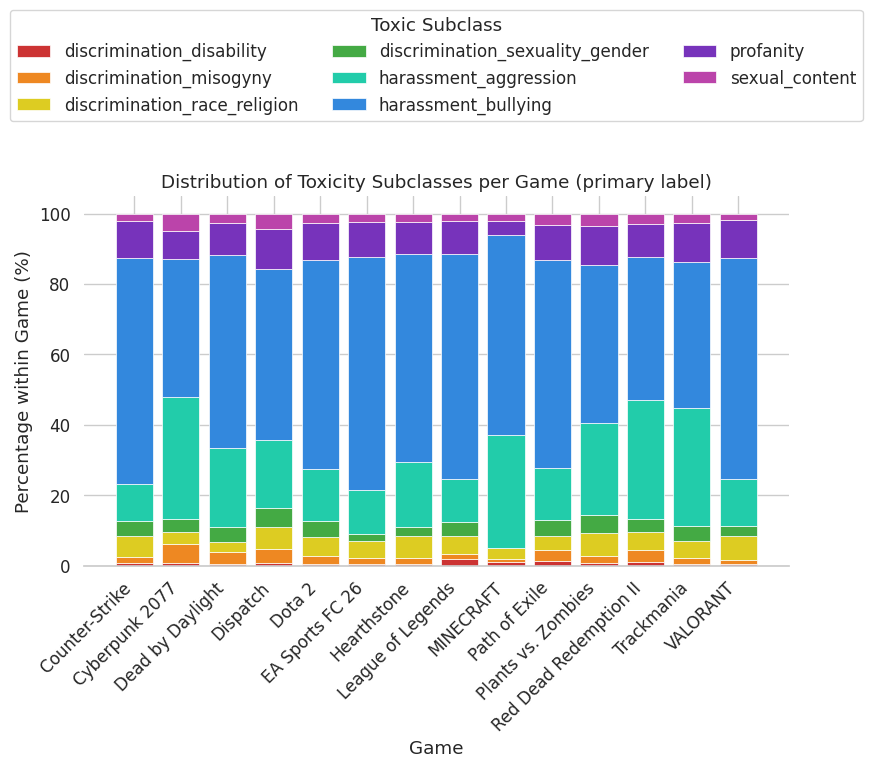}
    \includegraphics[clip, trim=0cm 0.2cm 0cm 5cm,width=0.98\columnwidth]{fig/game_subclasses.png}
    \caption{Distribution of toxicity subclasses per game.}
    \label{fig:toxic_cat_game}
\end{figure}

\begin{table}[t]
\centering
\caption{Relative amount of toxic chat messages per game.}
\label{tab:relative_toxicity_per_game}
\begin{tabular}{lr lr}
\toprule
\textbf{Game} & \textbf{Value} & \textbf{Game} & \textbf{Value} \\
\midrule
Red Dead Redemption II & 3.96 & Plants vs. Zombies & 2.37  \\
League of Legends & 3.34 & EA Sports FC 26 & 2.33  \\
Dispatch & 2.82 & Dead by Daylight & 2.21\\
Dota 2 & 2.75 & Hearthstone & 2.18\\
Path of Exile & 2.51 & Counter-Strike & 1.96 \\
Valorant & 2.48 & Trackmania & 1.38\\
Cyberpunk 2077 & 2.44 &  Minecraft & 0.74 \\
\bottomrule
\end{tabular}
\end{table}

\subsection{RQ3 Results: Variations of viewer toxicity between genres}

Multiplayer Shooter streams accounted for the largest share of messages (40.9\%), followed by MOBA (20.4\%), Sports Games (12.9\%), and Single-Player Shooter (1.5\%), while 24.3\% of messages came from games without an assigned genre. Toxicity rates varied by genre, with the highest proportion in MOBA streams (3.2\%), followed by Single-Player Shooter (3.1\%), and lower rates in Multiplayer Shooter (2.1\%) and Sports Games (2.0\%).

Table~\ref{tab:toxicity_by_genre_primary_secondary} shows the prevalence of primary and secondary toxicity labels among previously identified toxic messages by genre. Across all four genres, \textit{harassment} is the most frequent primary label category (74.0\%-77.6\%). \textit{Profanity}, in contrast, is comparatively rare as a primary label (8.8\%-10.2\%) but is the most frequent secondary label across all genres (50.5\%-55.0\%).

At the subclass level, \textit{bullying} is the most common primary harassment label in all genres (39.9\%-63.7\%), while \textit{aggression} is especially pronounced in Single Player Shooter streams (34.2\%). Within discrimination-related categories, Single Player Shooter also shows the highest prevalence of \textit{misogyny} in both the primary (4.4\%) and secondary labels (4.3\%). Overall, MOBA and Multiplayer Shooter exhibit very similar distributions across most primary and secondary categories, whereas Single Player Shooter differs more clearly through its elevated shares of \textit{aggression}, \textit{misogyny}, and \textit{sexual content}.

We compared per-stream distributions of toxicity subclasses in primary labels, using PERMANOVA~\cite{anderson2014permutational}. 
Pairwise comparisons showed significant differences between all genre pairs. However, PERMDISP~\cite{anderson2006distance} revealed unequal dispersion across all pairs, suggesting the PERMANOVA results may reflect differences in variability rather than distribution centers.
\begin{table*}[ht]
\caption{Primary and secondary toxicity label prevalence (\%) among identified toxic messages by genre. Overall rows report the combined prevalence for each top-level category. Bold indicates the highest prevalence per primary and secondary level within each row.}

\label{tab:toxicity_by_genre_primary_secondary}
\centering
\small
\resizebox{1.0\textwidth}{!}{
    \begin{tabular}{lcccccccc}
    \toprule
    & \multicolumn{4}{c}{Primary label (\%)} & \multicolumn{4}{c}{Secondary label (\%)} \\
    \cmidrule(lr){2-5} \cmidrule(lr){6-9}
    Category & MOBA & MP Shooter & SP Shooter & Sports Games & MOBA & MP Shooter & SP Shooter & Sports Games \\
    \midrule
    \textbf{Harassment (overall)}
      & 75.71 & 75.28 & 74.03 & \textbf{77.55}
      & \textbf{13.01} & 12.59 & 12.94 & 11.96 \\
    \quad Bullying
      & 62.84 & \textbf{63.69} & 39.89 & 59.29
      & \textbf{11.39} & 10.69 & 10.89 & 10.37 \\
    \quad Aggression
      & 12.87 & 11.58 & \textbf{34.15} & 18.25
      & 1.62 & 1.90 & \textbf{2.06} & 1.59 \\
    \addlinespace
    \textbf{Discrimination (overall)}
      & 12.40 & 12.14 & \textbf{13.30} & 9.67
      & 10.04 & 8.98 & \textbf{10.66} & 8.20 \\
    \quad Race / Religion
      & 5.12 & \textbf{6.32} & 4.24 & 4.77
      & 1.90 & 1.99 & 1.84 & \textbf{2.02} \\
    \quad Sexuality / Gender
      & \textbf{4.11} & 3.72 & 3.82 & 2.61
      & \textbf{2.73} & 2.41 & 2.48 & 1.99 \\
    \quad Misogyny
      & 1.66 & 1.40 & \textbf{4.44} & 1.81
      & 3.17 & 2.86 & \textbf{4.30} & 2.63 \\
    \quad Disability
      & \textbf{1.50} & 0.70 & 0.80 & 0.48
      & \textbf{2.24} & 1.73 & 2.04 & 1.58 \\
    \addlinespace
    \textbf{Sexual Content}
      & 2.27 & 1.95 & \textbf{3.87} & 2.56
      & 4.13 & 3.30 & \textbf{6.67} & 3.96 \\
    \textbf{Profanity}
      & 9.59 & \textbf{10.62} & 8.79 & 10.20
      & \textbf{54.99} & 52.00 & 50.53 & 52.27 \\
    \bottomrule
    \end{tabular}
}
\end{table*}

\section{Discussion}
\label{sec:discussion}
Our results show that the applied pipeline can successfully distinguish between toxic and non-toxic messages.
At the same time, the analyzed data only includes messages that remained visible in chat and were not removed by automatic or manual moderation. The observed prevalence of toxicity therefore already points to limitations of the existing moderation system, as harmful content is still present despite these filtering mechanisms.
Our proposed method achieves a level of agreement with human labeling, which is comparable to inter-human agreement and suggests that although toxicity definitions are subjective, we achieve similar reliability as human labelers.

Since chat messages are expected to be moderated, the comparatively high prevalence of toxicity in the dataset suggests a need for improved moderation mechanisms.
Twitch’s taxonomy is useful, but some subclasses are difficult to distinguish in practice. In particular, the distinction between harassment and profanity is often ambiguous, which is reflected in our agreement study. This suggests that the observed disagreement may not solely result from model error, but may instead point to limitations of the taxonomy itself.
Clearer category definitions could reduce annotation ambiguity and improve classification consistency, likely offering more practical value than subclass distinctions that cannot be applied reliably.

\subsection{RQ1: Commonly exhibited toxic behavior on Twitch}
Overall, the results indicate that while only a small part of messages on Twitch are toxic, most of them consist of insults and antagonizing language rather than explicit discriminatory or sexual content.
Harassment in the form of bullying or aggression makes up over half of the analyzed toxic messages, oftentimes accompanied by profanity, which commonly serves to intensify the tone of a message.
This goes in line with prior research~\cite{aguerri2023enemy}.
However, discriminatory messages still account for over 10\% of toxic messages, indicating that discrimination and hate speech remain persistent issues in online gaming communities. Although such messages occur less frequently, they are often perceived as particularly severe by players and can lead to disengagement from gaming spaces as well as negative impacts on mental well-being~\cite{fuchs2025players, fox2017women}. Since game-external communities serve as important spaces for player interaction and support, addressing toxicity on widely used platforms such as Twitch remains an important challenge~\cite{ma2024gaming}.

\begin{framed}
    \noindent\textbf{Finding 1}: Harassment is the most prevalent form of toxicity in Twitch messages and frequently co-occurs with profanity. Profanity most often appears as a secondary label, indicating that it is commonly used to intensify otherwise toxic language.
\end{framed}

\subsection{RQ2: Variations of viewer toxicity between games}

The analysis of toxicity ratios in specific games show that games differ not only in their overall toxicity rates but also in the composition of toxic behavior. This means that toxicity is not simply a platform-wide phenomenon but is shaped by the communities surrounding individual games.

A central result of our analysis is that almost all pairwise comparisons between games reveal significant differences in toxicity subclass distributions. This points toward the existence of game-specific community norms. Even when games are grouped under the same broader category, their chats do not necessarily develop the same toxicity profile. The clearest example is that some game pairs within a broader category remain distinct, whereas Counter-Strike~2 and Valorant appear similar. This similarity is plausible because both games share closely related competitive structures, audience expectations, and viewing practices, which may encourage similar forms of chat interaction.

The observed variation between games suggests that game mechanics, streamers, their communities, and the governing moderation process likely interact in shaping toxic behavior. Competitive games may foster more direct confrontation and blame, while games with different pacing, audience cultures, or forms of viewer engagement may produce different patterns of toxicity~\cite{dreier2023toxicity}.

For moderation, this implies that game- or community-specific adaptation may be more effective than one-size-fits-all approaches. If different games attract different mixtures of harassment, discriminatory language, sexualized remarks, and profanity, then moderation systems and community guidelines should be calibrated to the risks that are most characteristic of each game community.

\begin{framed}
    \noindent\textbf{Finding 2}: Harassment is the most prevalent form of toxicity across all games, often accompanied by profanity. While overall toxicity levels are similar between games, the distribution of toxicity categories differs significantly, except between Valorant and Counter-Strike~2. 
\end{framed}

\subsection{RQ3: Variations of viewer toxicity between genres}

Our results indicate that genre-related differences affect not only the occurrence of toxicity, but also which forms of toxicity are most prominent within each genre.
The ratios of toxic messages in genres show meaningful differences in overall toxicity rates, with MOBA streams displaying the highest relative toxicity and Sports Games the lowest among the analyzed genres. 
Comparing the distributions of toxicity subclasses shows that MOBA and Multiplayer Shooter chats appear particularly similar, whereas Single-Player Shooter stands out through comparatively higher shares of \textit{aggression}, \textit{misogyny}, and \textit{sexual content}. This suggests that genre (or the specific set of games selected per genre) can shape the prominence of certain toxic behaviors, but not necessarily the overall distribution of toxicity.

At the same time, the statistical results should be interpreted with caution. Although PERMANOVA indicates significant pairwise differences between all genre pairs, the corresponding PERMDISP results show unequal dispersion. This means that the observed genre differences may partly reflect different levels of within-genre variability rather than a clean separation between genre means.

Overall, this fits the broader pattern of the results: genre matters, but genre alone is not sufficient to explain the prevalence of toxicity categories. Considerable variation remains within most genres, although Multiplayer Shooters appear to be a partial exception, with comparatively similar profiles across games. This is consistent with the stronger role of individual games and communities observed in \textbf{RQ2}. Taken together, the results suggest that genre is a useful but limited analytical lens. It captures broad tendencies in toxicity prevalence and emphasis, but it does not fully account for the more fine-grained variation that emerges at the level of individual games and their surrounding communities.

\begin{framed}
    \noindent\textbf{Finding 3}:
    Genre influences both the overall level and the relative emphasis of toxic behavior on Twitch.
    Toxicity across genres shares a common underlying structure centered on harassment and profanity, while much of the finer-grained variation appears to be driven by individual games and their communities rather than genre alone.    
\end{framed}

\section{Limitations}
We discuss limitations of our work in alignment with the guidelines of Wohlin et al.~\cite{wohlin2012experimentation}. 
\paragraph{Construct Validity}
Toxicity is inherently subjective and context-dependent, and interpretations may vary across individuals and communities. Our labeling follows Twitch’s moderation taxonomy, though messages may still be interpreted differently depending on context or community norms. 
While human–model agreement is moderate, some false positives and negatives remain, but these rarely occur and are unlikely to affect aggregate patterns in the large-scale dataset.
\paragraph{Internal Validity}
The dataset primarily contains messages from popular streams, which may introduce selection bias. Communication dynamics and moderation practices can differ between large and smaller communities~\cite{dreier2023toxicity}. 
In addition, Twitch employs automated moderation tools that filter messages belonging to the categories described in Section~\ref{subsec:toxic_cat_selection}, meaning the dataset includes only messages that were not previously removed by the platform. 
Manual moderation by streamers and their moderation teams may further influence which messages remain visible in chat. As a result, the dataset may underestimate the absolute prevalence of toxic behavior, and observed differences in toxicity levels may partly reflect moderation practices rather than community behavior. 
However, the large scale of the dataset, spanning hundreds of streams and communities, reduces the likelihood that these factors systematically bias the overall results and still enables meaningful comparisons of relative toxicity patterns across games and genres.

To reduce confounding effects from differing gameplay mechanics and community structures, genre comparisons are limited to manually selected structurally similar games (e.g., Valorant and Counter-Strike 2), ensuring that comparisons remain meaningful.
Finally, non-English messages may be harder for the model to interpret and could not be evaluated by the human raters, potentially affecting reliability in multilingual contexts. However, due to our focus on English-speaking streamers, the vast majority of messages in the dataset are written in English, which limits the impact of this issue on the overall analysis.
\paragraph{Conclusion Validity}
Toxicity labeling was performed using a language model rather than a task-specific classifier, which may introduce classification inconsistencies.
To ensure consistency, we provide the model validation in Section~\ref{sec:model_agreement}.

\paragraph{External Validity}
Our findings are based on Twitch chat data and may therefore not fully generalize to other gaming communities or online platforms.
Communication on Twitch differs from many social media platforms due to its fast-paced chat and features such as emotes. However, as one of the largest gaming-related social platforms, it provides a valuable environment for studying toxicity in large gaming communities.
Our comparable results on the TextDetox dataset\footnote{\url{https://huggingface.co/textdetox/xlmr-large-toxicity-classifier-v2};\\\hphantom{ww}Last
 accessed: \today} further suggest a degree of generalizability.

\section{Conclusion}
\label{sec:conclusion}
This work examines toxicity in Twitch chats across games and genres using a large-scale dataset of more than 20 million messages and a zero-shot LLM-based classification pipeline. 

The proposed pipeline reliably distinguishes toxic from non-toxic messages, produces classifications comparable to human judgments, and performs well against the benchmark dataset, where it achieved strong results in comparison to the fine-tuned baseline model.
However, our findings also indicate that some limitations stem not only from the model, but from ambiguities in Twitch’s underlying toxicity taxonomy itself. 
Our results show that harassment is the dominant form of toxicity on Twitch and frequently co-occurs with profanity, while discriminatory and sexualized content occur less often but remain relevant.
We also find that toxicity varies not only in its overall prevalence, but also in its composition across games and genres. In particular, most games exhibit distinct toxicity profiles, suggesting that toxic behavior is shaped by game-specific contexts and community norms beyond broader genre-level effects.

Future work could refine the toxicity taxonomy, incorporate Twitch-specific features such as emotes, and apply user-based clustering to identify more detailed toxicity patterns across communities. Examining toxicity at the level of individual streams could provide a more fine-grained understanding of Twitch communities on their most fundamental level, improving the robustness of the classification approach.

\section*{Acknowledgements}
We are grateful for the support of Bodo Rosenhahn, which made the preparation and dissemination of this work possible.
Further, we disclose the use of LLMs for improving readability.

\bibliographystyle{IEEEtranS}
\bibliography{bib/references}

\newpage

\end{document}